\documentclass[journal]{IEEEtran}
\usepackage{hyperref}       
\usepackage{url}            
\usepackage{booktabs}       
\usepackage{amsfonts}       
\usepackage{nicefrac}       
\usepackage{microtype}      
\usepackage{graphicx}
\usepackage{floatrow}
\usepackage{bm}
\usepackage{amsmath}
\usepackage{color, xcolor}
\usepackage{floatrow}
\usepackage{cite}
\usepackage{soul}
\usepackage{amsthm}
\usepackage[linesnumbered,ruled,vlined]{algorithm2e}
\usepackage{verbatim} 
\newcommand{\xqedhere}[2]{%
	\rlap{\hbox to#1{\hfil\llap{\ensuremath{#2}}}}}

\begin{document}
%
\title{A Robust and Reliable Point Cloud Recognition Network Under Rigid Transformation}

\author{Dongrui Liu,~\IEEEmembership{Student Member,~IEEE}, Chuanchuan Chen,~\IEEEmembership{Student Member,~IEEE}, Changqing Xu,
        Qi Cai, Lei Chu,~\IEEEmembership{Member,~IEEE}, Fei Wen,~\IEEEmembership{Senior Member,~IEEE}, and Robert Qiu,~\IEEEmembership{Fellow,~IEEE}
\thanks{All authors are with Shanghai Key Laboratory of Navigation and Locationbased Services. School of Electronic Information and Electrical Engineering, Shanghai Jiao Tong University.}
}

\markboth{Journal}%
{Shell \MakeLowercase{\textit{et al.}}: Bare Demo of IEEEtran.cls for IEEE Journals}
%

\maketitle

\begin{abstract}
Point cloud recognition is an essential task in industrial robotics and autonomous driving. Recently, several point cloud processing models have achieved state-of-the-art performances. However, these methods lack rotation robustness, and their performances degrade severely under random rotations, failing to extend to real-world scenarios with varying orientations. To this end, we propose a method named Self Contour-based Transformation (SCT), which can be flexibly integrated into various existing point cloud recognition models against arbitrary rotations. SCT provides efficient rotation and translation invariance by introducing Contour-Aware Transformation (CAT), which linearly transforms Cartesian coordinates of points to translation and rotation-invariant representations. We prove that CAT is a rotation and translation-invariant transformation based on the theoretical analysis. Furthermore, the Frame Alignment module is proposed to enhance discriminative feature extraction by capturing contours and transforming self contour-based frames into intra-class frames. Extensive experimental results show that SCT outperforms the state-of-the-art approaches under arbitrary rotations in effectiveness and efficiency on synthetic and real-world benchmarks. Furthermore, the robustness and generality evaluations indicate that SCT is robust and is applicable to various point cloud processing models, which highlights the superiority of SCT in industrial applications.
\end{abstract}


%
\IEEEpeerreviewmaketitle

\section{Introduction}

%
%
%
%
\IEEEPARstart{T}{HREE} dimensional (3D) point clouds have attracted tremendous attention due to the requirement of modern applications, such as human-computer interaction \cite{lin2018recognition} and autonomous driving\cite{qiu2018rgb, wang2021mobile}. With the development of 3D sensors \cite{ li2021enhancing, lin2021long}, point clouds can be easily acquired and have been widely studied in robotics \cite{guo2014integrated, wang2019automatic}.

Due to irregularity and sparsity, traditional regular convolutional neural networks (CNNs) fail to handle point clouds directly. To take advantage of the powerful standard CNNs, point clouds are transformed into voxel-grids\cite{maturana2015voxnet}, mesh\cite{feng2019meshnet} and multi-view projections\cite{chen2017multi}, suffering from quantization artifacts and memory burden. PointNet\cite{qi2017pointnet} is the pioneer to consume point clouds directly with deep neural networks. After that, many similar methods emerge to various tasks, such as point cloud classification \cite{qi2017pointnet2, wang2019dynamic}, segmentation \cite{cui2020lightweight}, generation \cite{chen2021genecgan}, interpretation \cite{shen2021interpreting}, and detection \cite{liu2020high, xu2021object}.
 
  Although the advancement of computational resources and 3D sensors enables researchers to consume point clouds directly\cite{klokov2017escape, mao2019interpolated}, those methods are highly vulnerable to perturbations of random rotations. Unlike the synthetic and aligned dataset, orientations of raw point clouds generated by LiDAR sensors are typically unknown and dynamic. Efficient and precise classification and segmentation of point clouds against rotations are essential for real-world scenarios, \emph{e.g.}, autonomous driving. Thus, this paper mainly considers improving the rotation robustness of various existing point cloud processing models for classification and part segmentation.

A straightforward way to overcome the issue is to apply tremendous rotation augmentation to improve the orientation robustness of models. However, the infinite 3D rotation group (SO3) makes it impossible to design such a network with a high capacity to extract consistent shape awareness features against random rotations\cite{markley2014fundamentals}. Besides, it is computationally expensive and suffers from arbitrary rotation perturbations without meeting strict rotation invariance. Alternatively, several schemes have been recently proposed for developing rotation-invariant architectures\cite{you2018prin ,chen2019clusternet, zhang2019rotation, sun2019srinet, li2020rotation}. PRIN\cite{you2018prin} utilizes spherical voxel convolution to capture robust features without ensuring strict rotation invariance. ClusterNet\cite{chen2019clusternet}, RIConvNet\cite{zhang2019rotation}, and RI-Conv \cite{li2020rotation} build local descriptors to replace the Cartesian coordinates of points with relative angles and distances, which may cause the ambiguity of local shapes. SRINet\cite{sun2019srinet} encodes point clouds through a nonlinear mapping, inevitably impairing neighboring geometries. Obtaining rotation-invariant representations is essential for these methods, which calls for substantial analysis under the same backbone(s).

In this paper, we first compare and analyze existing rotation-invariant architectures, identifying rotation-invariant transformation methods that inhibit their extension to real-world industrial applications. Then we design a \textbf{Rotation and Translation-Invariant Transformation (RTIT)} module to enhance the rotation and translation robustness of various point cloud processing techniques via CAT. Furthermore, representations provided by CAT make neural networks more challenging to extract discriminative features, considering that each point cloud is transformed from the unified Cartesian frame to a self contour-based frame. To tackle this issue, we further introduce a \textbf{Frame Alignment (FA)} module to capture contours and transform its self contour-based frame to an intra-class frame.

\begin{figure*}[t]
	\centering
	\includegraphics[height=3cm, width=18cm]{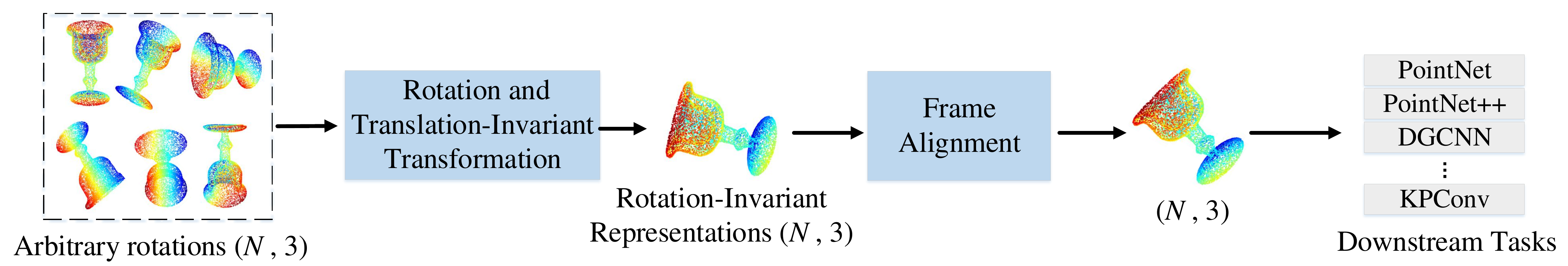}
	\caption{ Illustration of SCT architecture consisting of two novel modules: RTIT and FA. Point clouds are transformed by these two modules to obtain rotation-invariant representations and then fed into existing point cloud processing models for classification and segmentation tasks.}
	\label{fig:architecture}
\end{figure*}

Fig. \ref{fig:architecture} depicts the overall framework of SCT, consisting of RTIT and FA. The key contributions of this paper are summarized as follows:
\begin{itemize}

\item We introduce a novel RTIT, linearly transforming point clouds to the self contour-based rotation and translation-invariant representations while preserving global geometric structure. This new module can avoid the ambiguity of local shapes with low computational complexity.

\item The proposed SCT has good portability and can be flexibly integrated into various existing point cloud recognition models, \emph {e.g.}, PointNet \cite{qi2017pointnet} and DGCNN \cite{wang2019dynamic}.

\item Extensive experiments have been conducted on synthetic and real-world benchmark datasets for classification and part segmentation tasks to verify the generalization, efficiency, and robustness of the proposed method.

\end{itemize}

The rest of the paper is organized as follows. Section II introduces the related work. Section III illustrates the problem formulation. In Section IV, we elaborate the proposed framework is described, and provide mathematical analysis for the CAT. Furthermore, discussions on SCT are provided in Section V. Section VI illustrates experimental studies on synthetic and real-world datasets. Conclusions are presented in Section VII. 

\section{Related Work}
\label{gen_inst}
Many deep learning-based techniques have been proposed to process 3D point clouds in recent years. Recent learning-based techniques mainly include point cloud processing models and rotation-invariant models. Specifically, point cloud processing models consist of voxel-based, projection-based, and point-based models.

\subsection{Voxel and Probjection-based models.} To extend effective 2D convolution to 3D recognition, VoxNet\cite{maturana2015voxnet} and ShapeNets\cite{wu20153d} convert point clouds to volumetric representations and apply the standard 3D convolution. Even with modern GPUs, these techniques could only process low resolution voxel-grids (\emph{e.g.}, 32$\times$32$\times$32 in VoxNet\cite{maturana2015voxnet}). Octree\cite{riegler2017octnet}, Kd-Tree\cite{klokov2017escape} based methods were proposed to avoid the convolution in empty space to reduce memory consumption. OctNet\cite{riegler2017octnet} makes a prominent contribution, being able to handle high-resolution up to 256$\times$256$\times$256. 3DmFV \cite{ben20183dmfv} represents each point by a mixture of Gaussians and uses symmetric functions to compute global Fisher Vector representations \cite{sanchez2013image}. Along this direction, the following methods went on to process voxel and grids\cite{xu2020grid, wang2019octreenet}. Meanwhile, many works projected point clouds onto 2D images for recognition \cite{su2015multi, qi2016volumetric}. Though these organized and efficient data structures save time complexity, they inevitably induce higher memory cost and resolution loss.
\subsection{Point-Based models.} In another line, PointNet \cite{qi2017pointnet} is the pioneer to process the unordered point clouds directly with neural networks. It extracts point-wise features through shared Multilayer Perceptrons (MLPs) from the simply $(x,y,z)$ coordinates and adopts a symmetric function max pooling to get global features while pursuing permutation invariance. Since neglecting to mine neighbor relationship, various networks are proposed to remedy it. PointNet++ \cite{qi2017pointnet2} uses PointNets to hierarchically capture local features and enhance the local interactions. ECC \cite{simonovsky2017dynamic} analyzes neighboring points by spectral graph convolution and a graph pooling strategy. DGCNN \cite{wang2019dynamic} constructs dynamic local graphs and extracts semantic relation with EdgeConv operation. PointCNN \cite{li2018pointcnn} reorders local points with a convolution operation named $\chi$-Conv and achieves good performances on classification and part segmentation tasks. SO-Net \cite{li2018so} utilizes a self-organizing map and mini-PointNet to hierarchically establish neighbor interaction by extracting node-wise features. ShellNet \cite{zhang2019shellnet} uses statistics from spherical shells to define an efficient permutation invariant convolution operation for point cloud processing. KPConv\cite{thomas2019kpconv} proposes a kernel point convolution and deformable convolution with several kernel points for complex tasks. Besides, some works were proposed for point convolution \cite{wu2019pointconv, atzmon2018point, xu2018spidercnn}, kernel-based convolution\cite{lei2020spherical, komarichev2019cnn, liu2019relation}. However, most of them are vulnerable to random rotations, which are very general among real-world industrial applications.

\subsection{Rotation-Invariant models.} To apply these deep learning-based point cloud processing techniques to real-world industrial applications, some recent works explored rotation-invariant models. Thomas et al.\cite{thomas2018tensor} achieved local rotations equivalence by designing filters built from spherical harmonics and extending them to tensor field neural networks. The researchers\cite{esteves2018learning, cohen2018spherical} proposed a spherical convolution operation to learn rotation-invariant features from spherical representations. Similarly, PRIN \cite{you2018prin} employs Spherical Voxel Convolution to capture robust features. However, as these spherical voxel convolution-based techniques approximated the infinite group SO(3) with discretized rotation groups, they do not guarantee strict and global rotation invariance. ClusterNet\cite{chen2019clusternet}, RIConvNet\cite{zhang2019rotation}, and RI-Conv \cite{li2020rotation} build K-nearest neighbor graphs and transform Cartesian coordinates into relative angles and distances to harvest local rotation invariance. To extract expressive features, ClusterNet \cite{chen2019clusternet} utilizes a clustering operation, which is time-consuming and not applicable to the segmentation task. SRINet\cite{you2018prin} encodes point clouds through the cosine value of relative angles. RI-Conv \cite{li2020rotation} combines local and global features by two independent modules to yield a larger receptive field. However, the above methods encode relative information in the local patch, inevitably impairing geometric structures and causing ambiguity on flat surfaces. In contrast, the proposed CAT adopts a global linear transformation, which is rigorously rotation and translation-invariant while maintaining geometric structures.

\section{Problem Formulation}

The original point clouds collected by 3D sensors or laser scanners usually contain many attributes, such as 3D coordinates, RGB colors, surface normal, and intensity. A 3D point cloud with $N$ points is formulated as  ${\bm P} =( {\bm p}_{1},{ \bm p}_{2}, \ldots, {\bm p}_{N} )$ where ${\bm p}_{i} \in \mathbb R^{D}$ for $i= 1, 2, \ldots, N$, and $D$ denotes attributes. In this work, we only consider 3D coordinates, \emph{i.e.}, $D=3$. Point cloud data have two main distinct properties:

\begin{itemize}

\item \textbf{Unorganized}. As a collection of 3D point coordinates, point clouds are different from structured images or volumetric grids, without the specific order. Thus, with any permutation operations, point clouds still stand the same object, maintaining geometry topology.

\item \textbf{Rotation and Translation Invariance}. Point clouds are discrete representations of continuous surfaces of 3D space, indicating that rotations and translations should not change the semantic category of the object nor the intrinsic structures. Therefore, point cloud processing models need rotation and translation robustness.
\end{itemize}

For point cloud recognition tasks, the distribution of 3D point clouds needs to be modeled. Specifically, we adopt a deep network to learn the high-dimensional latent representations of point clouds that preserve the original point clouds' geometry structures and semantic features for further applications, \emph{e.g.}, classification and segmentation. Learning latent representations of point clouds is also called a feature extraction process.

 \textbf{Without perturbations.} 
 The process of extracting latent representations of point clouds without perturbations is formulated as
\begin{equation}
{ F} = \Phi ({ P}) \in \mathbb{R}^{C},
\end{equation}
where $\Phi: \mathbb R^{3\times N} \to \mathbb R^{C}$ denotes a Feature Extraction Network, and ${ F}$ is a $C$-dimensional vector descriptor, reflecting the original point clouds' semantic features. Then through a classifier $f$, we can get an output label,
\begin{equation}
{l} = f ({ F}) \in \mathbb{R}^{L},
\end{equation}
where $l$ denotes the predicted category label of the original point cloud object.

\begin{figure}
	\includegraphics[height=4cm, width=6cm]{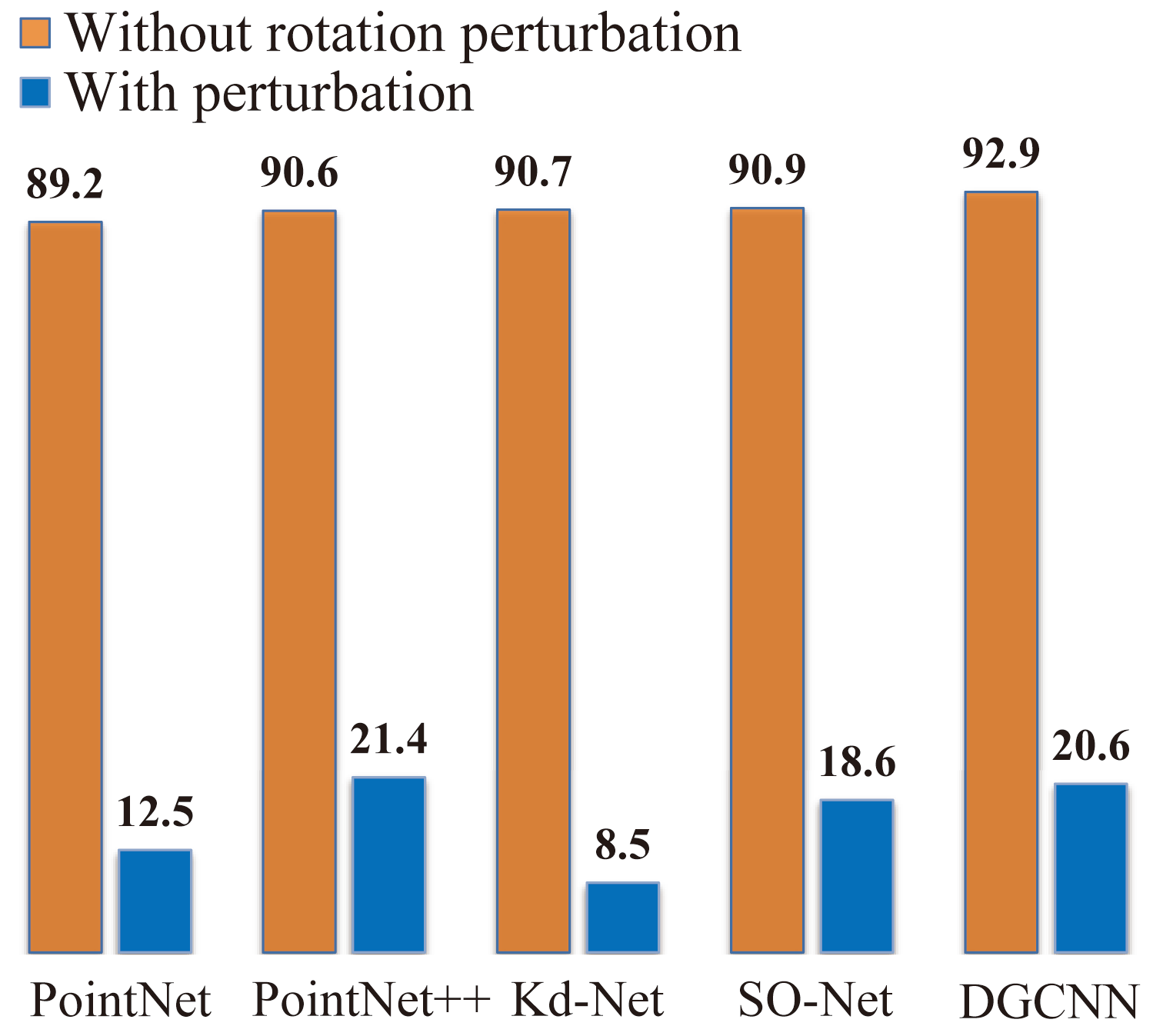}
	\caption{Classification accuracy (\%) on ModelNet40 with or without rotation perturbations. It is clear that the performances of existing point cloud processing models drop sharply under rotation. More comparison experiments are presented in Section \ref{Experiments}.}
	\label{fig:comparison}
\end{figure}

\par \textbf{With perturbations.} Many recent proposed state-of-the-art techniques have achieved more than 92\% accuracy on ModelNet40\cite{wu20153d}, depending on effective feature extraction networks. The process of feature extraction under perturbations of rotation and translation is formulated as:
\begin{equation}
{\ F} = \Phi ({ RP+T}) \in \mathbb{R}^{C},
\label{567}
\end{equation}
 where ${R}\in SO(3)$ represents a rotation matrix and ${T}$ denotes a translation matrix. However, the existing feature extraction models fail to learn rotation-invariant shape-awareness. As shown in Fig. \ref{fig:comparison}, their performances degrade severely under random rotations, lacking rotation robustness. 
 
 To enhance rotation and translation robustness of feature extraction network $\Phi$, we introduce the concept of RTIT. Given a point cloud ${ P} =({\bm p}_{1},{ \bm p}_{2}, \ldots, { \bm p}_{N})$, a strict RTIT can be expressed as a mapping $G$ such that
 \begin{equation}
 G{ (RP+T) }=  G({ P}),
 \end{equation}
 where ${ R}\in SO(3)$ denotes a rotation matrix and ${ T} = ({\bm t},{ \bm t}, \ldots, {\bm t})$ is a $3 \times N $ translation matrix with $ {\bm t} \in \mathbb R^{3}$. As shown in Fig. \ref{fig:comparison}, the feature extraction process of neural networks is sensitive to rotation and translation perturbations. Thus, it is reasonable to find a proper $G$ to eliminate effects of rotation and translation transformations for a given point set $P$ without information loss. Accordingly, we reformulate the process of feature extraction in Eq. (\ref{567}) with RTIT, as follows, 
 \begin{equation}
 {\ F} = \Phi ({ G(P)}) \in \mathbb{R}^{C}.
 \label{579}
 \end{equation}
 
 Eq. (\ref{579}) shows the rotation and translation robustness of feature extraction network $\Phi$, i.e., $\Phi ({ G(RP+T)})=\Phi ({ G(P)})$.

 In this paper, as illustrated in Fig. \ref{fig:architecture}, we propose SCT, which handles the arbitrary orientation point clouds and maps them to rotation and translation-invariant representations. It is noted that SCT can be directly incorporated by various existing start-of-the-art point cloud recognition methods for classification and segmentation tasks while keeping rotation and translation invariance. 

\section{Method and Analysis}
\label{section3}

In this section, we introduce the proposed SCT, which consists of two modules:
\begin{itemize}
	
	\item[1)] Rotation and Translation-Invariant Transformation. An RTIT module transforms Cartesian coordinates of points to rotation and translation-invariant representations. 
	
	\item [2)] Frame Alignment. A Frame Alignment (FA) module enhances point clouds by detecting contour points and regresses the quaternion. The quaternion is transformed to a coordinate system alignment matrix, aiming to transform each object from its self contour-based frame to an intra-class frame where discriminative features are easier to extract.
\end{itemize}
Equipped with RTIT and FA, rotation-sensitive point cloud processing models (\emph{e.g.}, PointNet) are robust to rotation and translation, thus can be applied to real-world scenarios.

\begin{figure}[t]
	\centering
	\includegraphics[width=0.99\linewidth]{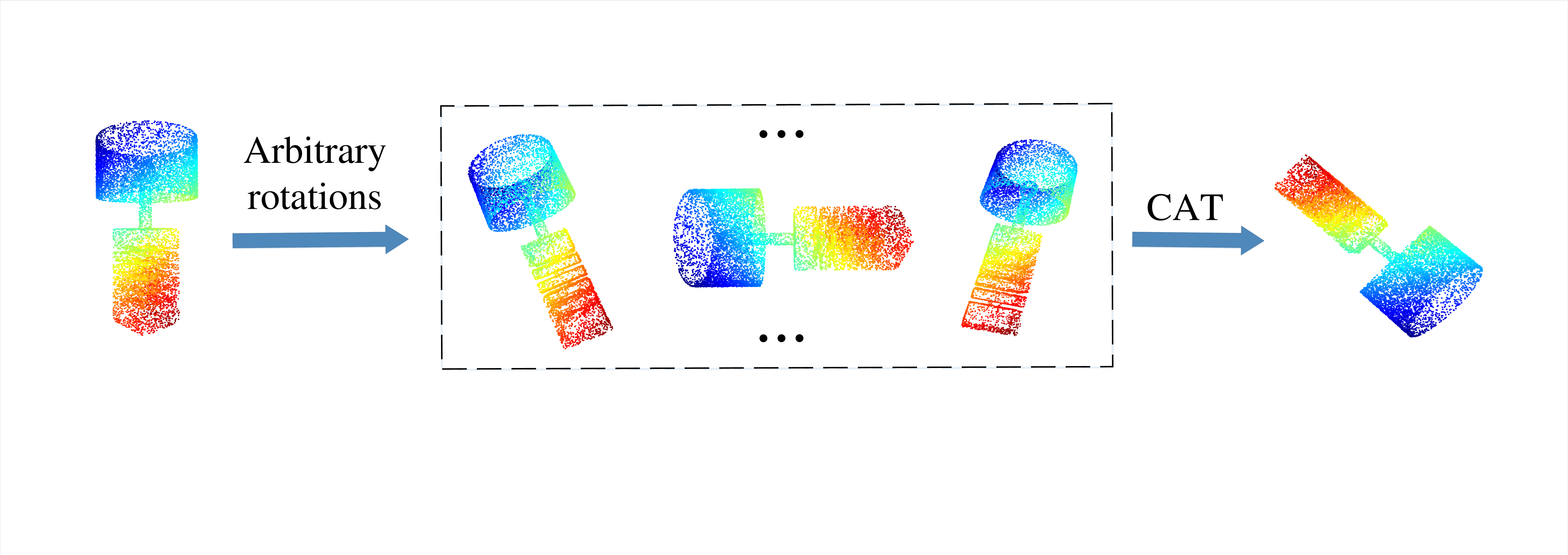}
	\caption{Intuitive explanations on the effect of CAT.}
	\label{fig:cat_effect}
	\vspace{-0.2cm}
\end{figure}

\subsection{Design of A Linear RTIT}
\label{RTIT}
The concept of RTIT has been defined in the above section. Here, we aim to propose a linear RTIT called \textbf{Contour-Aware Transformation} (CAT). ClusterNet \cite{chen2019clusternet} is a nonlinear RTIT, which converts the independent coordinates to relative distance-based representations through a nonlinear transformation to harvest rotation and translation invariance. Different from the linear transformation, the nonlinear transformation disturbs the data distribution inevitably. Inspired by works on point set registration\cite{yang2015go, wu2020rigid} that linearly align two rigid transformed point sets, we aim to eliminate effects of rotation and translation by a simple yet effective CAT.

We first calculate the barycenter of points ${\bm p}_{b}:=\frac{1}{n}\sum_{i=1}^{n}{\bm p}_{i}\\=(x_{b},y_{b},z_{b})^{T}$. In this way, ${P_{b}}=({\bm p}_{b}, \ldots, {\bm p}_{b})$ is a $3 \times N$ matrix. The farthest and closest point from barycenter are derived as:
\begin{align}
{\bm p}_{f}=(x_{f},y_{f},z_{f})^{T} = \arg \max_{{\bm p}_{i}\in P} \| {\bm p}_{i} - {\bm p}_{b}\|_{2}, \nonumber \\
{\bm p}_{c}=(x_{c},y_{c},z_{c})^{T}= \arg \min_{{\bm p}_{i}\in P} \| {\bm p}_{i} - {\bm p}_{b}\|_{2}.
\end{align}
Then, we define vector ${\bm \beta}_{f}$, ${\bm \beta}_{c}$ and ${\bm \beta}_{n}$, respectively, where ${\bm \beta}_{f}:={\bm p}_{f}-{\bm p}_{b}$, ${\bm \beta}_{c}:={\bm p}_{c}-{\bm p}_{b}$, $\bm\beta_{n}:=\bm \beta_{c} \times \bm \beta_{f}$, and $\times$ is cross product. To construct orthogonal axes, the vector $\bm \beta_{c}$ is finally updated, $\bm \beta_{c} = \bm \beta_{f} \times \bm \beta_{n}$. 

We generate a new self contour-aware frame $B=[{\bm\beta_{f}, \bm\beta_{n}, \bm\beta_{c}}]$. A normalized and orthogonal frame is derived as
\begin{equation}
B=[{\bm\beta_{f}/\|\bm\beta_{f}\|, \bm\beta_{n}/\|\bm\beta_{n}\|, \bm\beta_{c}/\|\bm\beta_{c}}\|]=[{ X^{'}\ Y^{'}\ Z^{'}}],
\label{5}
\end{equation}
and the point set $P$ is accordingly transformed into
\begin{equation}
P' =G_{CAT}(P)= B^{T}(P-P_{b}).
\label{730}
\end{equation}

Eq. \ref{730} defines CAT, which aims to find three self contour-aware axes and linearly transforms the frame to harvest rotation and translation-invariant representations (See Proposition 1 for theoretical analysis). We further give an intuitive explanation of the effect of CAT as shown in Fig. \ref{fig:cat_effect}. In this way, CAT can eliminate effects of rotation transformations and obtain rotation invariance.

\par Parameter-free CAT is agnostic to the feature extraction network. Thus CAT could be integrated into various point cloud processing models. Detailed experimental results in Section \ref{Experiments} demonstrate its efficiency and effectiveness for synthetic and real benchmarks.

\begin{figure}[t]
	\centering
	\includegraphics[width=0.98\linewidth]{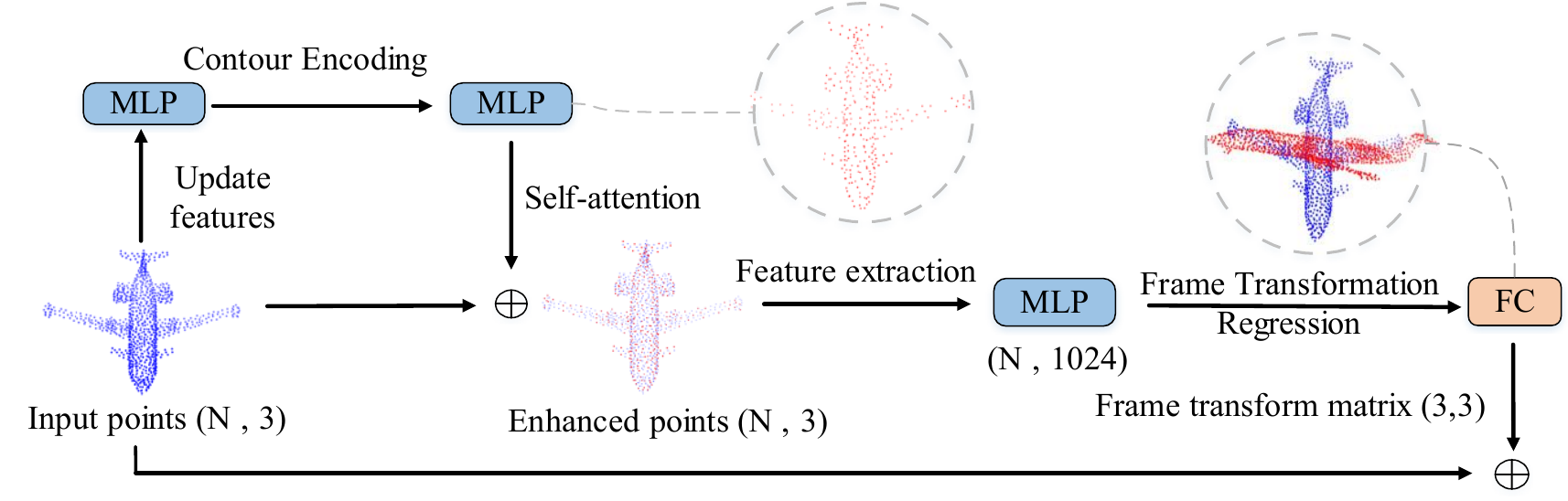}
	\caption{Architecture of FA. FA consists of Contour Encoding and Frame Transformation Regression.}
	\label{fig:rotation_net3}
\end{figure}

\begin{figure*}[t]
	\includegraphics[width=0.95\linewidth]{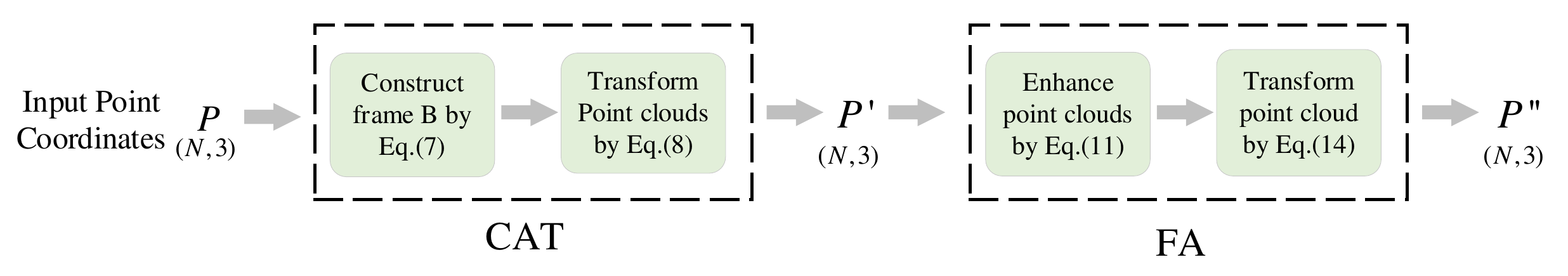}
	\caption{The flowchart of the technological process of the proposed SCT.}
	\label{fig:flowchart}
	\vspace{-0.3cm}
\end{figure*}

\vspace{-0.3cm}
\subsection{On the Novel FA Module}
\label{fa}
The CAT transforms points from the Cartesian Coordinate to each self contour-based frame to obtain rotation and translation invariance. However, the feature extraction network has to learn discriminative features across different frames instead of the unified Cartesian Coordinate, which is much more difficult. To address the problem, for each rotation and translation-invariant representation provided by RTIT, we leverage a FA module to capture contours and transform its self contour-based frame to an intra-class frame to enhance discriminative feature extraction. Intra-class frame aims to find a frame that discriminates one class from others (\emph{e.g.}, desks and monitors have different intra-class frames for better discrimination). As shown in Fig. \ref{fig:rotation_net3}, FA consists of two modules: Contour Encoding and Frame Transformation Regression.
\paragraph{Contour Encoding}
Given rotation and translation-invariant representations $ P'$, this unit explicitly captures contours and augments these points through a self-attention mechanism, which explores the geometric structure and eventually benefits the entire optimization process. To this end, the operation is formulated as:
\begin{eqnarray}
f_{c} = \sigma(\phi({ P'})), { P'} \in \mathbb R^{N\times3}, \\
{ P'_{c}} = softmax(\phi(f_{c})), { P'_{c}} \in \mathbb R^{N\times3},
\end{eqnarray}
where $\sigma$ is a nonlinear activator, and $\phi$ is chosen as an MLP. The first MLP maps point clouds from a three-dimension space to a high-dimensional space and harvests latent representations. Then we employ another MLP to reduce feature dimensions to 3D, preserving salient features. Finally, a softmax function is used to figure out key points. We enhance the rotation and translation-invariant representations $P'$ with contour features $P'_{c}$, resulting in enhanced representations $P'_{a}$,
\begin{equation}
 P'_{a} = P' \oplus P'_{c},
\end{equation}
where $\oplus$ denotes channel-wise summation.
\paragraph{Frame Transformation Regression}
The Frame Transformation Regression transforms point clouds from its self contour-based frame to an intra-class frame. The input of Frame Transformation Regression is the point cloud enhanced by Contour Encoding module. The architecture follows the encoder-decoder framework. For the encoder, we adopt a simple PointNet-like\cite{qi2017pointnet} structure, resulting in global features $f_{global}$. Afterward, several fully connected layers are used as a decoder to regress the frame transformation quaternion. Mathematically, the pipeline of FA is formulated as
\begin{align}
f_{global} = \mathcal {A}(\gamma ({ P'_{a}})) \in\mathbb R^{C},\\
[q_{0}, q_{1}, q_{2}, q_{3]}]= \theta (f_{global}) \in\mathbb R^{4},
\end{align}
where ${ P'_{a}}$ are points enhanced by Contour Encoding; $\gamma$ denotes PointNet-like\cite{qi2017pointnet} 1024-dimensional vector descriptor; $\mathcal A$ stands aggregation function, \emph{i.e.}, max-pooling; and $\theta$ is fully connected layers. The transformation matrix $T_{est}$ can be computed as
\begin{equation}
\begin{aligned}
T_{est} &= {
	\left[ \begin{array}{ccc}
	1-2q_{2}^{2}-2q_{3}^{2} & 2q_{1}q_{2}-2q_{3}q_{0} & 2q_{1}q_{3}+2q_{2}q_{0}\\
	2q_{1}q_{2}+2q_{3}q_{0} & 1-2q_{1}^{2}-2q_{3}^{2} & 2q_{2}q_{3}-2q_{1}q_{0}\\
	2q_{1}q_{3}-2q_{2}q_{0} & 2q_{2}q_{3}+2q_{1}q_{0} & 1-2q_{1}^{2}-2q_{2}^{2}
	\end{array} \nonumber
	\right ]}. \\
\end{aligned}
\end{equation}

Finally, the point cloud $P''$ transformed by FA is obtained as
\begin{equation}
P'' = T_{est}P' \in \mathbb{R}^{3 \times N}.
\label{859}
\end{equation} 
The point cloud $P''$ can be flexibly fed to various point cloud recognition models for classification and segmentation tasks with rotation and translation robustness. 

\subsection{Training Classification and Segmentation Models with SCT}

Here, we present detailed procedures of training classification and segmentation models with SCT. Firstly, given a point cloud $ P \in \mathbb{R}^{3 \times N}$, CAT transforms point cloud $P$ into $P'\in \mathbb{R}^{3 \times N}$ to obtain rotation and translation-invariant representations. Secondly, FA enhances the point cloud $P'$ and obtains point cloud $P'' \in \mathbb{R}^{3 \times N}$ to ease the feature extraction process. In this way, the enhanced point cloud can be directly fed into existing point cloud processing models (\emph{e.g.}, PointNet) for classification and segmentation tasks. Specifically, the point cloud segmentation task is a per-point classification task \cite{qi2017pointnet}, \emph{i.e.}, the segmentation task is to assign a label to each point in a 3D scan.  Fig. \ref{fig:flowchart} summarizes the detailed technological process of the proposed SCT. Together with the illustration in Fig. \ref{fig:architecture} and technical explanation, one can arrive at the whole point cloud recognition procedure (or other downstream tasks, e.g., Classification and Segmentation, etc.).  

\subsection{Theoretical Analysis of CAT}
Before proving CAT is an RTIT, we revisit the rotation via a new perspective based on the adjoint map \cite{markley2014fundamentals}, which is given in Lemma 1 as below:
\\
{\bf Lemma 1.} \textit{For $R \in SO(3)$, 
	\begin{equation}
	[(R{\bm x })\times] = R[{\bm x}\times]R^{T}, \nonumber
	\end{equation}
	where  ${\bm x}=(x_{1},x_{2},x_{3})^{T} $ and $[\bm x \times]$ is the cross product matrix, defined by
	\begin{equation}
	[\bm x \times]={\left[ \begin{array}{ccc}
		0 & -x_{3} & x_{2}\\
		x_{3} & 0 & -x_{1}\\
		-x_{2} & x_{1} & 0
		\end{array} 
		\right ]}. \nonumber
	\end{equation}}
The proof is given in Appendix. The adjoint map formulated in Lemma 1 enlightens us to exploit the property of cross product under rotation transformation. Lemma 2 is given below based on Lemma 1.\\
\\
{\bf Lemma 2.} \textit{For $R \in SO(3)$, 
	\begin{equation}
	(R {\bm \alpha_{1}})\times(R{\bm \alpha_{2}})=R( {\bm \alpha_{1}}\times{\bm \alpha_{2}}), \nonumber
	\end{equation}
	where ${\bm \alpha_{1}}=(x_{1},y_{1},z_{1})^{T}, {\bm \alpha_{2}}=(x_{2},y_{2},z_{2})^{T}$.}\\
\textit{Proof.} According to Lemma 1,
\begin{align}
(R{\bm \alpha_{1}})\times(R{\bm \alpha_{2}}) = &[(R{\bm \alpha_{1}})\times]R{\bm \alpha_{2}}  \nonumber \\ 
=& R[{\bm \alpha_{1}}\times]R^{T}R{\bm \alpha_{2}} \nonumber \\
=& R({\bm \alpha_{1}}\times{\bm \alpha_{2}})
\label{4}
\end{align}

Lemma 2 provides the consistency under rotation, i.e., if two vectors ${\bm \alpha_{1}}$ and ${\bm \alpha_{2}}$ are transformed by a rotation matrix $R$, whose cross product vector ${\bm \alpha_{1}}\times{\bm \alpha_{2}}$ will be transformed by the same rotation matrix consistently. Therefore, it is possible to eliminate effects of rotation via a dedicated linear transformation. The following theoretical analysis explains the favorable rotation-invariant property of CAT, which is beneficial for point cloud data processing under arbitrary rotations. \\

{\bf Proposition 1.} \textit{CAT mapping $G_{CAT}:P \to P'$ satisfies the definition of RTIT such that  $P'=G_{CAT}(P)=G_{CAT}(RP+T)$ holds for any translation matrix $ T\in \mathbb R^{3\times N}$ and any rotation mapping $R\in SO(3)$.}\\
\textit{Proof.}
Let $\overline{P}=RP+T = (\overline{\bm p}_{1},\overline{\bm p}_{2}, \ldots, \overline{\bm p}_{N})$, \\then we have
\begin{equation}
\begin{aligned}
\overline{B} &= ({\overline {\bm \beta}}_{f}, {\overline {\bm \beta}}_{n}, {\overline {\bm \beta}}_{c}) \label{12} \\
&= ({\overline {\bm \beta}}_{f}, {\overline {\bm \beta}}_{c} \times {\overline {\bm \beta}}_{f}, {\overline {\bm \beta}}_{f} \times ({\overline {\bm \beta}}_{c} \times {\overline {\bm \beta}}_{f})),
\end{aligned}
\end{equation}
and,
\begin{align}
{\overline {\bm \beta}}_{f}&={\overline{\bm p}}_{f}-{\overline{\bm p}}_{b}= R({\bm p}_{f}-{\bm p}_{b})=R{{\bm \beta}}_{f} \label{11} , \\
{\overline {\bm \beta}}_{n}&={\overline {\bm \beta}}_{c} \times {\overline {\bm \beta}}_{f} =R{{\bm \beta}}_{c} \times R{{\bm \beta}}_{f} \label{8}.
\end{align}
Considering Eq. (\ref{4}) (Lemma 2), we have 
\begin{align}
R{{\bm \beta}}_{c} \times R{{\bm \beta}}_{f} = R({{\bm \beta}}_{c} \times { {\bm \beta}}_{f})=R{{\bm \beta}}_{n}.
\end{align}
Eq. (\ref{8}) reduces to
\begin{align}
{\overline {\bm \beta}}_{n} = R{{\bm \beta}}_{n}.
\label{9}
\end{align}
Similarly, we have
\begin{align}
{\overline {\bm \beta}}_{c}= {\overline {\bm \beta}}_{f} \times ({\overline {\bm \beta}}_{c} \times {\overline {\bm \beta}}_{f}))=R[{{\bm \beta}}_{f} \times ({ {\bm \beta}}_{c} \times { {\bm \beta}}_{f})]=R{{\bm \beta}}_{c}.
\label{10}
\end{align}
Substituting Eq. (\ref{11}), Eq. (\ref{9}), and Eq. (\ref{10}) into Eq. (\ref{12}) yields
\begin{align}
\overline{B}= ({\overline {\bm \beta}}_{f}, {\overline {\bm \beta}}_{n}, {\overline {\bm \beta}}_{c})=R({{\bm \beta}}_{f}, {{\bm \beta}}_{n}, {{\bm \beta}}_{c})=RB.
\end{align} Thus, new representations with the novel $G_{CAT}$ mapping are
\begin{equation}
\begin{aligned}
\overline{P}'  & =G_{CAT}({\overline P})\\
&= {\overline B}^{T}({\overline P}-{\overline P}_{b}) \\
&= B^{T}R^{T}[RP-T-RP_{b}+T)]  \\
&= B^{T}(P-P_{b}). \xqedhere{118.5pt}{\qed}
\end{aligned}
\end{equation}

In addition to the above analysis, we will provide more detailed  discussion of the novel SCT in the following section. All these will contribute to a better understanding of the proposed methods' mechanism.

\section{Discussions on SCT}
\label{section4}
In this section, we give discussions and complexity analysis of SCT\footnote{It is noted that here we test and compare these rotation-invariant transformation methods on ModelNet40 \cite{wu20153d}.}.

\subsection{Discussion on the Relation between PCA and CAT}
\label{selecting}

Here, we aim to explain the intuition behind the proposed strategy. RTIT is similar to pose normalization, which aims to align an object into a canonical coordinate frame \cite{fu2008upright}. Principal Component Analysis (PCA) is the most commonly used method to solve the problem by transforming point coordinates to a new frame composed of three eigenvectors \cite{duda2006pattern}. However, PCA does not satisfy RTIT rigorously \cite{fu2008upright, yu2020deep}, i.e., the orientation candidate set of objects reduces to eight candidates. Note that the intuition is to figure out an RTIT rather than reduce the orientation candidate set. In addition, Table \ref{selecting_vector} shows that PCA normalization is rotation-variant.

\setlength{\tabcolsep}{10pt} 
\begin{table}[t]
	\begin{center}
		\caption{Comparison of CAT and PCA.}
		\label{selecting_vector}
		\begin{tabular}{cccc}
			\toprule
			Method  & NR/NR  & NR/AR  &$\bigtriangleup$Acc\\
			\midrule
			PCA+ DGCNN &  88.8  & 76.5  & -12.3 \\
			CAT+ DGCNN & \bf 89.0 & \bf 89.0  & 0 \\
			\bottomrule
		\end{tabular}
	\end{center}
\vspace{-0.6cm}
\end{table}

\vspace{-0.3cm}

\setlength{\tabcolsep}{14pt}
\begin{table}[ht]
	\begin{center}
		\caption{\textbf{Performance comparisons of rotation invariant transformation methods on ModelNet40.} $^{*}$ denotes our methods.}
		\label{comparison1}
		\begin{tabular}{ccc}
			\toprule
			Method     & Accuracy &Type  \\
			\midrule
			SRINet\cite{sun2019srinet}+ DGCNN  & 86.1 &NT\\
			RIConvNet\cite{zhang2019rotation}+ DGCNN  & 86.1 &NT\\
			RI-Conv\cite{li2020rotation}+ DGCNN  & 82.6 &NT\\
			ClusterNet\cite{chen2019clusternet}+ DGCNN & 86.4 &NT  \\
			\hline
			CAT$^{*}$ + DGCNN & \bf 89.0 &LT  \\
			\bottomrule
		\end{tabular}
	\end{center}
\vspace{-0.5cm}
\end{table}

\setlength{\tabcolsep}{7pt} 
\begin{table}[ht]
	\begin{center}
		\caption{\textbf{Comparison of computational complexity.} $^{*}$ denotes our methods. NT and LT represents
			nonlinear and linear transformation, respectively.}
		\label{comparison2}
		\begin{tabular}{ccc}
			\toprule
			Method     & Computational complexity&Type  \\
			\midrule
			SRINet\cite{sun2019srinet} & $O(N)$ &NT\\
			RIConvNet\cite{zhang2019rotation} & $O(NK)$ &NT  \\
			RI-Conv\cite{li2020rotation}  & $O(NK)$ &NT\\
			ClusterNet\cite{chen2019clusternet} & $O(NK)$ &NT  \\
			\hline
			CAT$^{*}$  & $O(N)$ &LT  \\
			\bottomrule
		\end{tabular}
	\end{center}
\vspace{-0.4cm}
\end{table}

\subsection{Discussion on the Type of RTIT}
\label{type}
We classify existing RTIT methods into two types, linear and nonlinear methods. Specifically, we compare and analyze their performances and computational complexity. 
\begin{itemize}
	
\item \textbf{Nonlinear Transformation} consists of ClusterNet\cite{chen2019clusternet}, SRINet\cite{you2018prin}, RIConvNet\cite{zhang2019rotation}, and RI-Conv \cite{li2020rotation}. SRINet\cite{you2018prin} chooses three axes and maps each point into a collection of relative angles and leverages a key point detection module to improve performance. ClusterNet\cite{chen2019clusternet}, RIConvNet\cite{zhang2019rotation} and RI-Conv \cite{li2020rotation} build K-nearest neighbor graphs and transform Cartesian coordinates into relative angles and norms to obtain rotation invariance. Overall, nonlinear RTIT methods convert point coordinates into relative angles and distances.

\item \textbf{Linear Transformation} includes the proposed CAT, which preserves geometry structures, avoiding introducing ambiguity in the process of nonlinear transformation.

\end{itemize}
To show the effectiveness of the proposed RTIT, we test and compare these rotation-invariant transformation methods on ModelNet40\cite{wu20153d}. As these compared approaches modify network architectures, we choose DGCNN\cite{wang2019dynamic} as the backbone of them for a fair comparison. We uniformly sample 1024 points from CAD models. Only the $(x,y,z)$ coordinates of sampled points are used. The qualitative and quantitative evaluation results are summarized in Table \ref{comparison1} and Table \ref{comparison2}, where NT and LT represent nonlinear and linear transformation, respectively.
 
\begin{figure}[t]
	\centering
	\includegraphics[height=4cm, width=7.5cm]{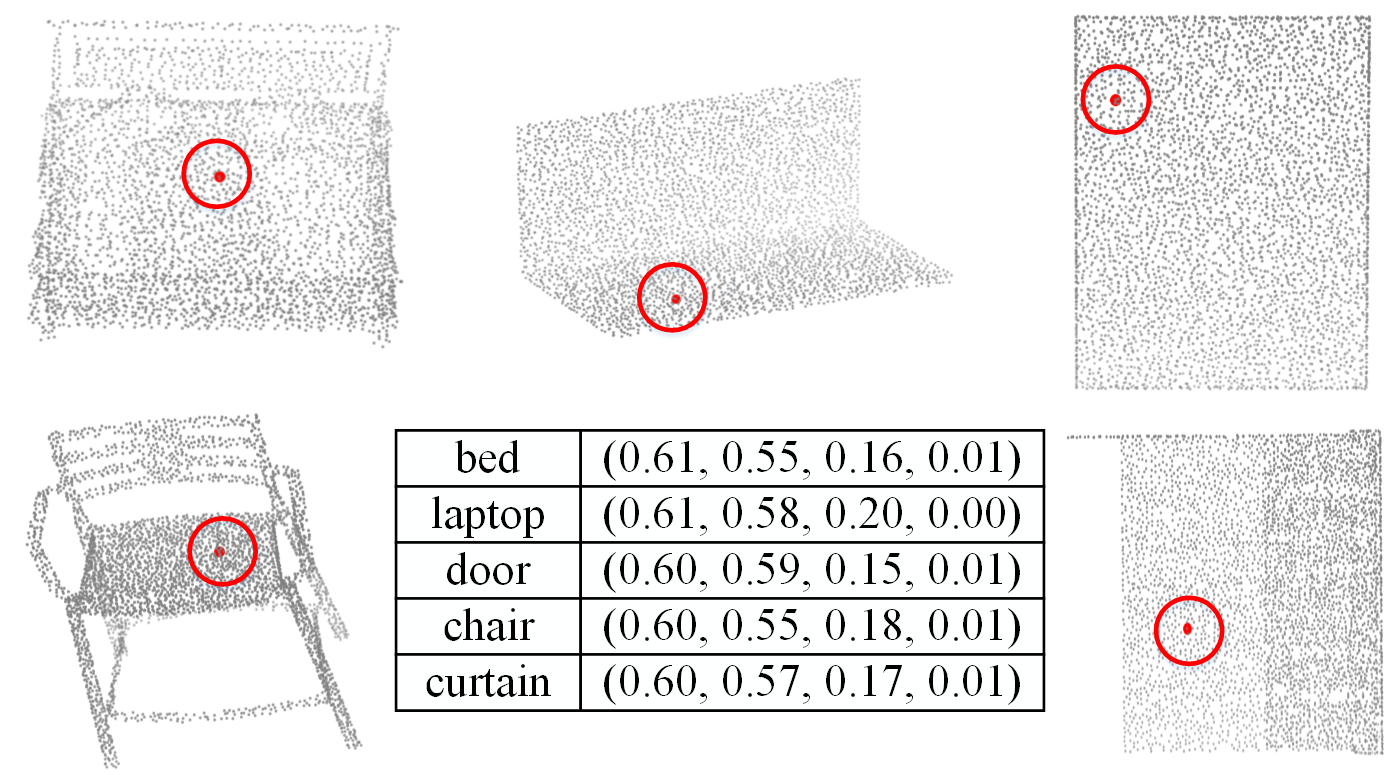}
	\caption{Ambiguity of local shapes in nonlinear transformation. Red points denote the center of local patch. The transformed representations of each point are shown in the table.}
	\label{fig:Ambiguity}
\end{figure}

\begin{figure*}
\centering
\includegraphics[height=5cm, width=17cm]{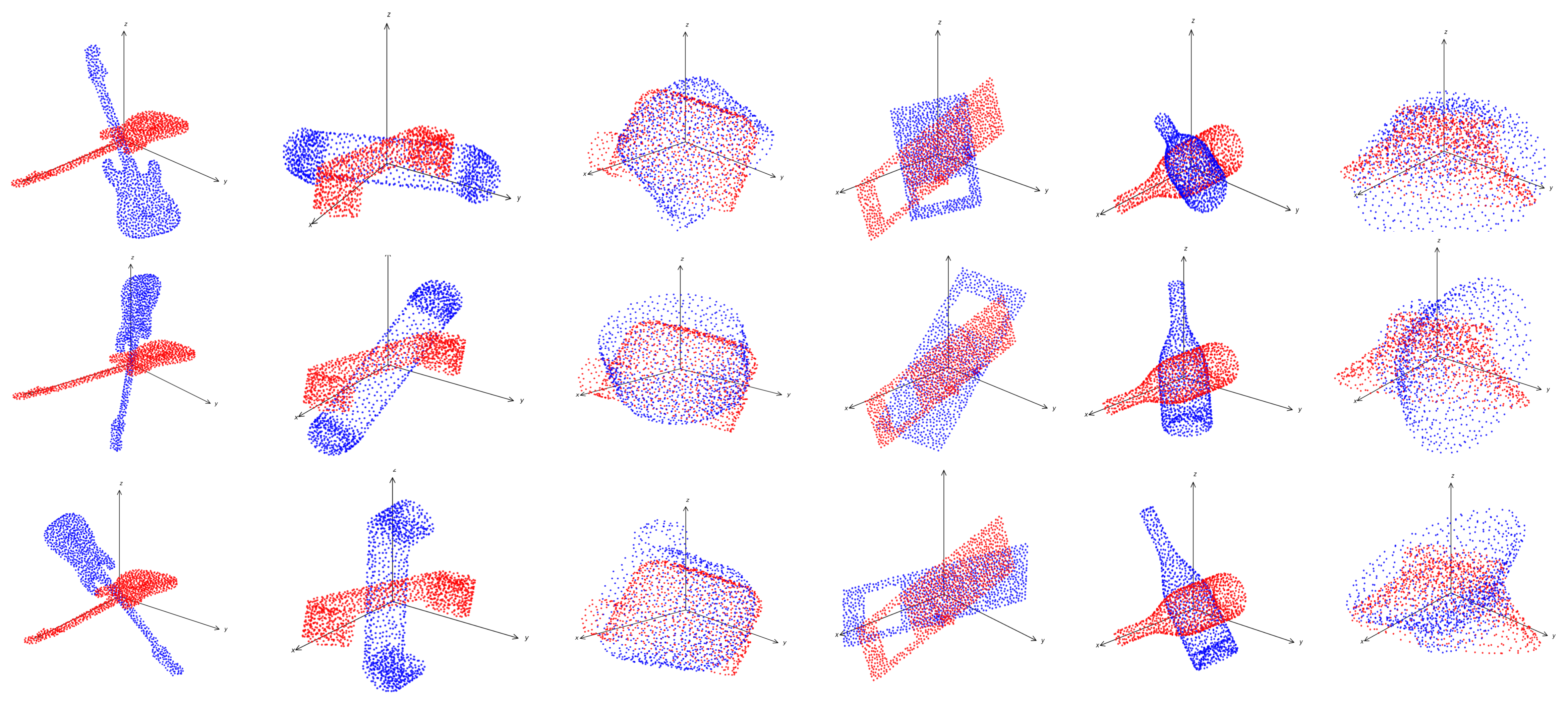}
\caption{Visualization of CAT. Original 3D objects and transformed by CAT are blue and red, respectively. Each column demonstrates the same object with different rotation perturbation.}
\label{fig:rit}
\end{figure*}

Though ClusterNet, RIConvNet, and RI-Conv build local graphs and encode relative angles, they still suffer from the loss of geometry information. We give a quantitative example to illustrate the ambiguity of local shapes caused by the nonlinear transformation. As shown in Fig. \ref{fig:Ambiguity}, the local patch (in red circles) on the flat surfaces belong to different objects. However, if we represent the center points of these flat local patches by the features transformed by ClusterNet \cite{chen2019clusternet}, the features belonging to the parts of various objects may be the same to each other. Therefore, similar features of different objects cause the ambiguity of local shapes. These ambiguous representations hinder networks from extracting features from distinguishing different shapes. Compared to these nonlinear transformations, CAT preserves the geometry structures by a linear transformation. Visualization examples in Fig. \ref{fig:rit} show that CAT consistently transforms objects into rotation-invariant representations (i.e., the effect of rotation is eliminated.). 
\par Based on the above analysis, CAT leads the classification performance under arbitrary rotation. In addition, CAT has lower computational complexity than nonlinear methods. Therefore, linear CAT outperforms other nonlinear methods and is more advantageous for industrial applications.

\subsection{Discussion on the FA Module}
The aim here is to demonstrate that FA is a key part of our architecture. RTIT module generates rotation and translation-invariant representations at the cost of transforming the Cartesian frame to a self contour-based frame. To alleviate the issue, we need a coordinate system alignment matrix to transform each object from its self contour-based frame to an intra-class frame. T-Net proposed in PointNet\cite{qi2017pointnet} predicts an affine transformation matrix for feature alignment. However, the nonlinearly affine transformation harms geometry relation and suffers from information loss inevitably. Thus, it is not applicable to this work. Furthermore, equipped with T-Net, PointNet is still vulnerable to random rotations. For coordinate frame alignment, the transformation matrix should be a rotation matrix. Thus, the FA module is introduced. 
\par
\begin{figure*}
\vspace{-0.3cm}
\centering
\includegraphics[height=7cm, width=17cm]{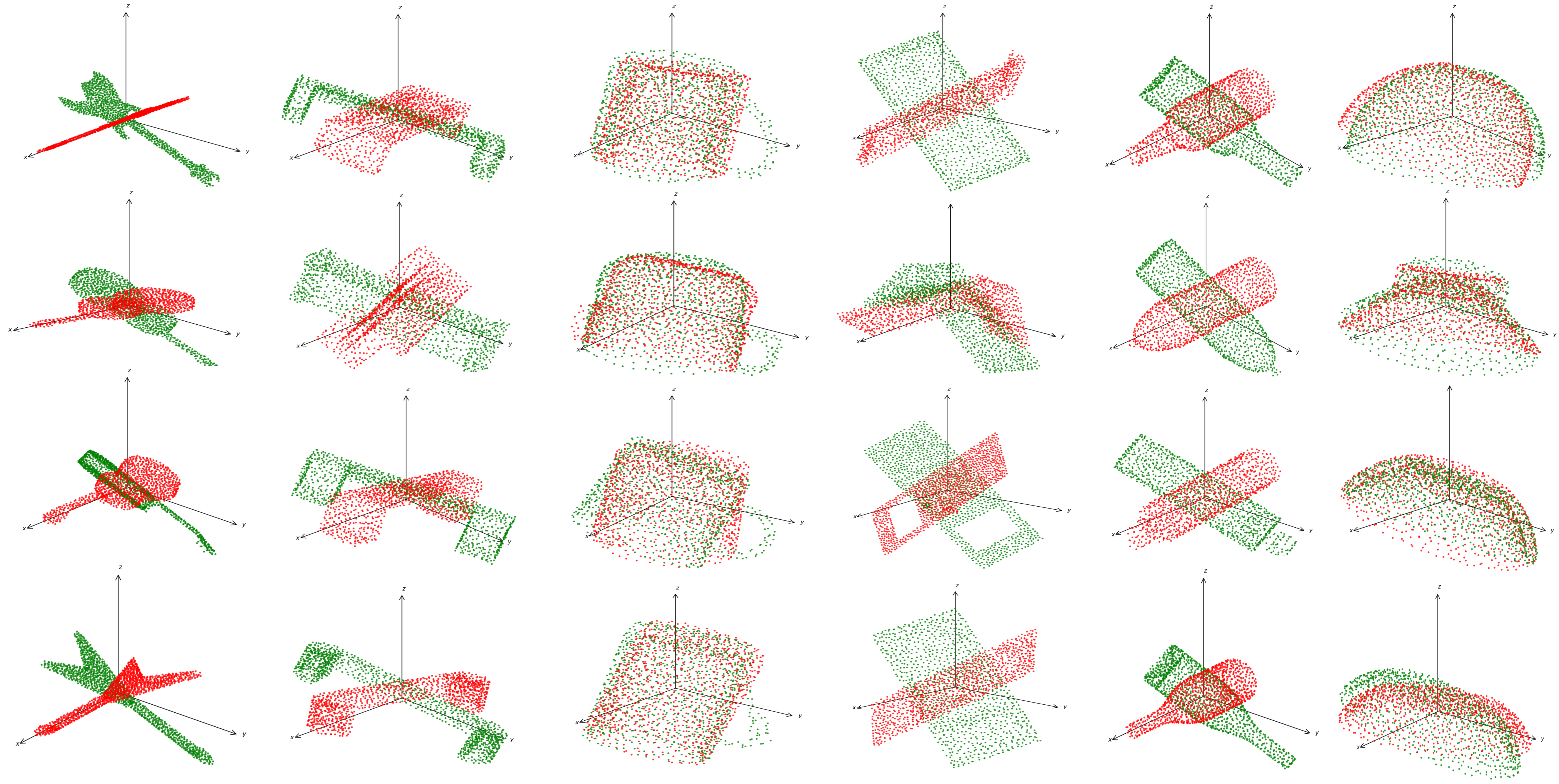}
\caption{Visualization of FA. Rotation and translation-invariant representations generated by CAT and 3D objects transformed by Frame Aligment are red and green, respectively.}
\label{fig:fa}
\end{figure*}

 Point clouds transformed by FA are visualized in Fig. \ref{fig:fa}. The input and output of FA are red and green, respectively. In Fig. \ref{fig:fa}, each column illustrates four different objects belonging to the same category. The transformed point clouds from FA in the same category have a similar intra-class frame (pose). For example, the legs of different benches have similar orientations, and the handles are on the same side of the cups. This phenomenon demonstrates the effectiveness of FA.

\setlength{\tabcolsep}{9pt} 
\begin{table}
	\begin{center}
		\caption{Classification results on ModelNet40.$*$ denotes training with azimuthal rotation data augmentation.}
		\label{classification}
		\begin{tabular}{cccc}
			\toprule
			Method     & NR/NR  & NR/AR   & AR/AR  \\
			\midrule
			SubVolSup \cite{qi2016volumetric}  & 88.5 & 36.6  & 82.7\\
			SubVolSup MO \cite{qi2016volumetric}  & 89.5 & 45.5 & 85.0 \\
			MVCNN 12x \cite{su2015multi}$^{*}$  & 89.5 & 70.1  & 77.6\\ 
			MVCNN 80x \cite{su2015multi}$^{*}$  & 90.2 & 81.5 & 86.0 \\ 
			PointNet\cite{qi2017pointnet}  & 89.2  & 12.5  & 80.3\\
			Kd-Net\cite{klokov2017escape}  & 90.7   & 8.5 & 79.2 \\
			PointNet++\cite{qi2017pointnet2}  & 90.6 & 21.4 & 85.0\\
			DGCNN\cite{wang2019dynamic}& \bf 92.9 & 20.6 & 81.1\\
			PointCNN \cite{li2018pointcnn}  & 92.2 & 29.6  & 84.5 \\
			KPConv \cite{thomas2019kpconv}  & 90.7 & 28.8 & 83.6\\
			\hline
			PRIN\cite{you2018prin}  & 80.1 & 70.4&-\\
			RIConvNet\cite{zhang2019rotation}  & 86.4 & 86.4 & 86.4 \\
			ClusterNet\cite{chen2019clusternet}  & 87.1 & 87.1 & 87.1 \\
			SRINet\cite{sun2019srinet}  & 87.1 & 87.1 & 87.1 \\
			\hline
			Our & 89.5    & \bf 89.5 & \bf 89.5 \\
			\bottomrule
		\end{tabular}
	\end{center}
	\vspace{-0.5cm}
\end{table}

\section{Experiments}
\label{Experiments}
In this section, we conduct extensive experiments on several point cloud classification and part segmentation benchmark datasets including ModleNet40\cite{wu20153d}, ScanObjectNN\cite{uy2019revisiting}, and ShapeNet\cite{yi2016scalable}. Some experimental results are visualized to demonstrate the effectiveness of the proposed technique. The point clouds transformed by SCT can be easily processed by existing point cloud recognition models while maintaining rotation invariance, which improves the rotation robustness of various existing models. We further feed these transformed point clouds into the DGCNN\cite{wang2019dynamic} model for classification and part segmentation tasks under arbitrary rotations and conduct an ablation study. All the experiments are implemented on two NVIDIA TITAN Xp GPUs in a distributed manner. We train and test models with three different settings to compare with state-of-the-art point cloud recognition architectures under arbitrary rotations. 1) Models are trained and tested without rotation (NR/NR). 2) Models are trained without rotation augmentation and tested with arbitrary rotations (NR/AR). 3) Arbitrary rotations are added during both the training and testing process (AR/AR).

\setlength{\tabcolsep}{9pt} 
\begin{table}[t]
	\begin{center}
		\caption{Classification results on ScanObjectNN.}
		\label{classification2}
		\begin{tabular}{ccccc}
			\toprule
			Method     & NR/NR  & NR/AR &AR/AR \\
			\midrule
			PointNet\cite{qi2017pointnet} &  79.8  & 24.9  & 70.4 \\
			PointNet++\cite{qi2017pointnet2}& 85.5 & 26.9  & 81.5 \\
			DGCNN\cite{wang2019dynamic}  & 86.2 & 27.2   & 78.7  \\
			PointCNN\cite{wang2019dynamic}  &  86.3 & 29.6   & 70.8 \\
			KPConv \cite{thomas2019kpconv} &\bf 88.8 & 44.4 & 78.5  \\
			\hline
			RIConvNet\cite{zhang2019rotation} &  73.5 & 73.5  &73.5 \\
			ClusterNet\cite{chen2019clusternet}&  80.4 & 80.4  & 80.4 \\
			SRINet\cite{sun2019srinet}& 78.8 & 78.8  & 78.8\\
			\hline
			Our  &84.6     & \bf 84.6 & \bf 84.6 \\
			\bottomrule
		\end{tabular}
	\end{center}
	\vspace{-0.3cm}
\end{table}

\begin{table*}[t]
	\caption{Part segmentation results on ShapeNet dataset. Metric is mean IoU(\%).}
	\label{part_segmentation}
	\setlength{\tabcolsep}{9pt}
	\centering
	\begin{tabular}{cccccccc}
		\toprule
		Method& Rotation-Invariant &Input &NR/NR &NR/AR & $\bigtriangleup mIoU_1$& AR/AR & $\bigtriangleup mIoU_2$\\
		\midrule
		PointNet\cite{qi2017pointnet}&No& 2048$\times $3 &83.2 & 31.3&-51.9 & 74.4&-8.8\\
		PointNet++\cite{qi2017pointnet2}& No& 2048$\times $3 &84.6 &36.7&-47.9& 76.7 &-7.9\\
		PointCNN\cite{li2018pointcnn}& No& 2048$\times $3 & \bf 84.8& 27.3&-57.5 & 71.4& -13.4\\
		SpiderCNN\cite{xu2018spidercnn}& No& 2048$\times $3 & 82.4& 35.5&-46.9 & 72.3&-10.1\\
		DGCNN\cite{wang2019dynamic}&No& 2048$\times $3 &84.7& 43.8 &-40.9& 73.3&-11.4\\
		ShellNet\cite{zhang2019shellnet}&No& 2048$\times $3 &82.8& 40.8 &-42.0& 77.1&-5.7\\
		\hline
		PRIN\cite{you2018prin}&Yes& 2048$\times $3 &71.5& 57.4 &-14.1& 68.9 &-2.6 \\
		RIConvNet\cite{zhang2019rotation}&Yes & 2048$\times $3 & 75.5 & 75.3 &-0.2 &75.5& 0 \\
		SRINet\cite{sun2019srinet}&Yes& 2048$\times $3 &77.0& 77.0 &0& 77.0&0\\
		RI-Conv\cite{li2020rotation}& Yes & 2048$\times $3 & 79.2 & 79.2 & 0 & 79.4 & 0 \\
		\hline
		Our &Yes& 2048$\times $3 &81.4& \bf 81.4 &\bf 0& \bf 81.4& \bf 0\\
		\bottomrule
	\end{tabular}
	\vspace{-0.3cm}
\end{table*}

\begin{figure*}[t]
\setlength{\abovecaptionskip}{-0.4cm}
\centering
\includegraphics[height=5.3cm, width=17cm]{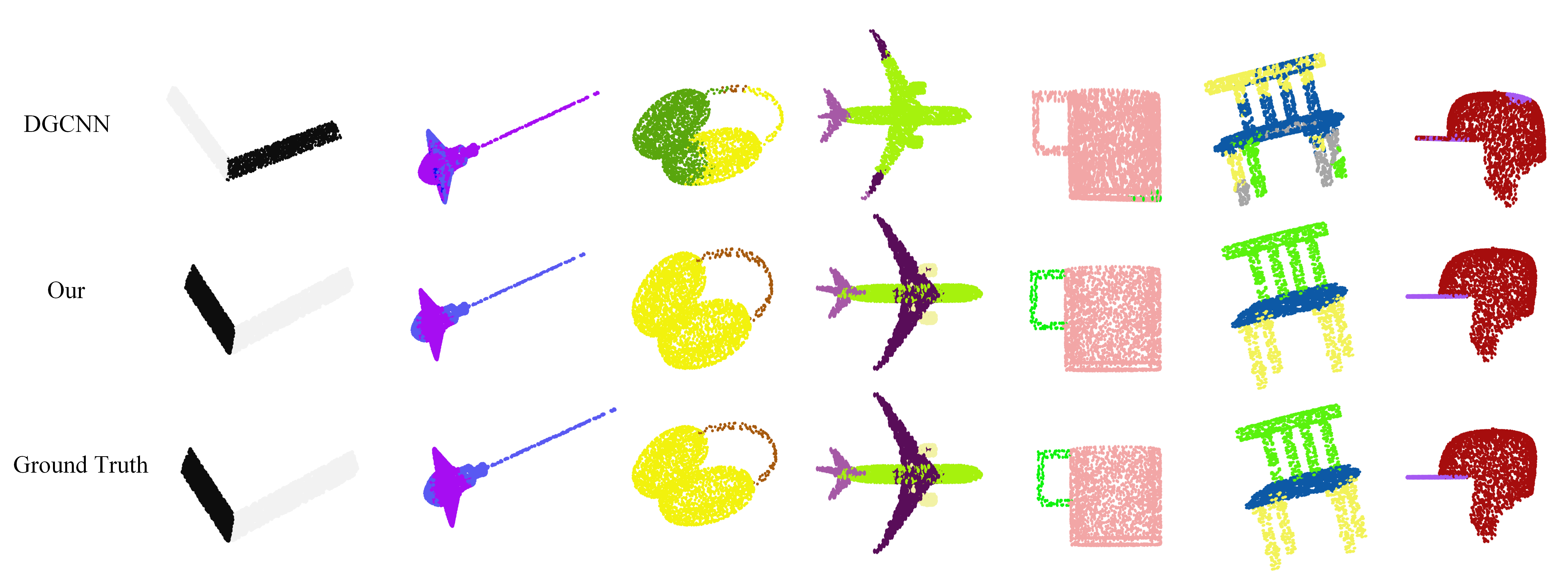}
\caption{Visualization of part segmentation results on ShapeNet under arbitrary rotations.}
\label{fig:shapenet}
\end{figure*}

\vspace{-0.3cm}
\subsection{Synthetic ModelNet40}
We conduct the classification tasks on the ModelNet40\cite{wu20153d} dataset under arbitrary rotations, consisting of 40 different categories with 9843 synthetic CAD training models and 2468 testing models. We uniformly sample 1024 points from models' surfaces to train our model. Following the experimental settings of PointNet\cite{qi2017pointnet}, these sampled point clouds are translated and rescaled into a unit sphere later. As 3D sensors could not capture normals directly in the real world, only the $(x,y,z)$ coordinates of sampled points are used as the input of models. During the training process, random scaling and jittering are added to perturb points' original positions. The training strategy is almost the same as \cite{wang2019dynamic} except that training epochs are changed to 290. \\
 \textbf{Evaluation} Table \ref{classification} shows results without voting trick. The proposed method achieves the best accuracy under random rotations on the ModelNet40 dataset. Though the current point cloud recognition methods have harvested very high accuracy, their performances degrade severely with rotations perturbation, implying the vulnerability to rotations. PointNet and DGCNN drop more than 70\% classification accuracy under rotations, failing to generalize to arbitrary orientations. Though training with arbitrary rotations improves the rotation robustness, the infinite 3D rotation group (SO3) makes it impossible for existing models with high capacity to extract consistent shape awareness features against random rotations. Therefore, under AR/AR setting, there still exists a performance drop from NR/NR. Compared with other rotation-invariant methods, our model is 2.3\% better than ClusterNet\cite{chen2019clusternet} and SRINet\cite{you2018prin}. The proposed method has a stable classification performance with/without rotation augmentation, proving its rotation invariance.

\subsection{Real-world ScanObjectNN}
Different from synthetic benchmark ModelNet40\cite{wu20153d} dataset, ScanObjectNN\cite{uy2019revisiting} is a newly published real-world dataset comprising of 2902 3D objects in 15 categories. We implement the classification task on the ScanObjectNN\cite{uy2019revisiting} to further prove the robustness and generalization ability of our technique in real-world scenarios. The training strategy, network architecture, and input settings are the same as the synthetic benchmark. All experiments are conducted in ``object onl'' of data split 1.\\
 \textbf{Evaluation} Note that SRINet\cite{you2018prin} needs normal vectors as input, while ScanObjectNN does not contain normal attributes. Thus, we report the performance of SRINet in Table \ref{classification2} without normal vectors. Compared with the synthetic dataset, the real-world benchmark is more challenging, with a noticeable performance drop for all the compared methods. Table \ref{classification2} shows that the proposed method outperforms the others under random rotations. The consistent performances on synthetic and real-world datasets show the generalization and robustness of our model, demonstrating its potential applications in the industrial point cloud processing tasks with arbitrary rotations.

\subsection{ShapeNet}
In this section, we evaluate our method for part segmentation task on ShapeNet part\cite{yi2016scalable} dataset containing 16,881 3D objects from 16 categories and 50 annotated parts in total. Part segmentation is a fine-grained classification task aiming to assign a semantic label to each point of a 3D object. Each object contains less than 6 part category labels. We randomly sample 2048 points from each object and split the dataset into train, validation, and test parts as the official scheme.\\ 
 \textbf{Evaluation} For a fair comparison, we use the mean Intersection-over-Union (mIoU) metric proposed in PointNet. Results of the proposed method and other techniques are shown in Table \ref{part_segmentation}. Note that ClusterNet\cite{chen2019clusternet} is dedicated to classification and not applicable to the part segmentation application. Those state-of-the-art models lack rotation robustness, failing to classify each part of the object with sharp performances drop under rotations (Fig. \ref{fig:shapenet}). In our experiments, training with rotation augmentation gives PointNet++ and DGCNN an approximately 20\% boost in mIoU, achieving 58\% and 62.7\%, respectively. Though rotation augmentation slightly improves their rotation robustness, there is still a performance gap between those models and our method. As discussed in Section \ref{type}, nonlinear transformation methods, such as RIConvNet\cite{zhang2019rotation}, SRINet\cite{sun2019srinet} and RI-Conv\cite{li2020rotation}, inevitably cause the ambiguity of local shapes which hamper the network to extract discriminative shape awareness. The comparison of classification results shown in Table \ref{part_segmentation} demonstrates that the proposed linear method outperforms other nonlinear methods, achieving 81.4\% mIoU.

 \begin{figure*}[t]
	\centering
	\includegraphics[width=0.95\linewidth]{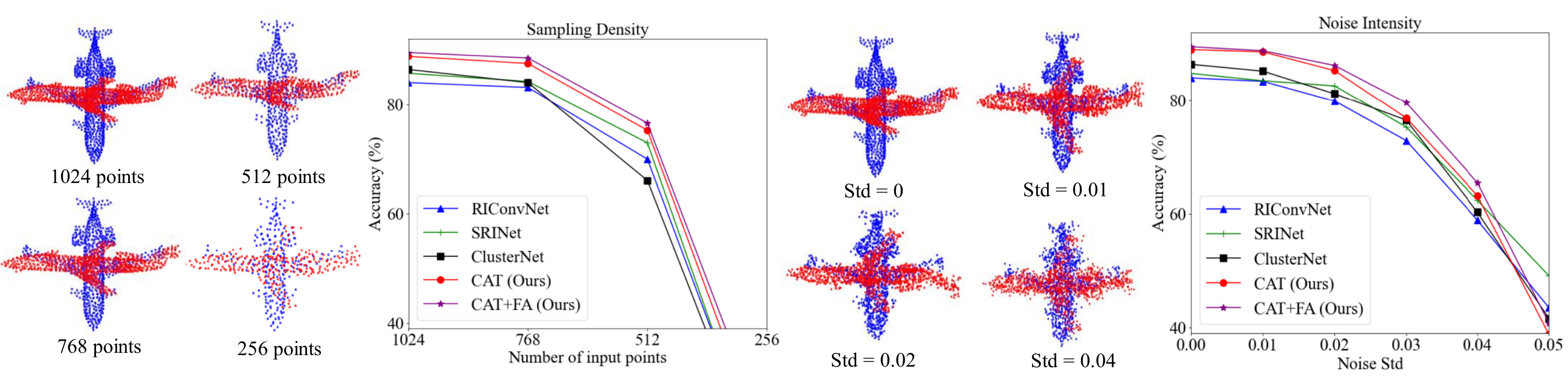}
	\caption{Robustness evaluation. Rotation-Invariant Transformation methods under different sampling densities and noises. Blue stands original points and red denotes transformed points by CAT approach. All the methods are evaluated in NR/AR.}
	\label{fig:sample1}
\end{figure*}

\setlength{\tabcolsep}{10pt} 
\begin{table}[t]
	\begin{center}
		\caption{RTIT integrated into various point cloud processing models. $^{*}$ denotes our methods.}
		\label{versatility}
		\setlength{\tabcolsep}{9pt}
		\begin{tabular}{cccc}
			\toprule
			Method  & NR/NR  & NR/AR  &$\bigtriangleup$Acc\\
			\midrule
			PointNet\cite{qi2017pointnet} & 89.2  & 12.5  & -76.7 \\
			PointNet++\cite{qi2017pointnet2}& 90.6 & 21.4  & -69.2 \\
			PointCNN\cite{li2018so} & 92.2  & 29.6 & -63.6\\
			DGCNN\cite{wang2019dynamic} & \bf 92.9 & 20.6  & -72.3 \\
			\hline
			CAT$^{*}$ + PointNet & 83.5 & 83.5 & 0 \\
			CAT$^{*}$ + PointNet++ & 88.3 & 88.3 & 0 \\
			CAT$^{*}$ + PointCNN  & 86.1 & 86.1 & 0\\
			CAT$^{*}$ + DGCNN & 89.0 & \bf 89.0 & 0 \\
			\bottomrule
		\end{tabular}
	\end{center}
\vspace{-0.5cm}
\end{table}

\setlength{\tabcolsep}{1.5pt}

\subsection{Generality of RTIT}
In this section, we use CAT to enhance the rotation and translation robustness of four classical networks, such as PointNet \cite{qi2017pointnet}, PointNet++ \cite{qi2017pointnet2}, PointCNN \cite{li2018pointcnn}, and DGCNN \cite{wang2019dynamic}. As discussed in Section \ref{section3}, CAT is agnostic to the feature extraction network. CAT maps the arbitrary orientation point cloud to rotation and translation-invariant representations, which can be directly processed by various point cloud recognition methods. Table \ref{versatility} indicates that equipping with parameter-free CAT enables these classical point cloud processing models to obtain rotation robustness, \emph{i.e.}, without performance degradation under rotation perturbations, which also demonstrates the effectiveness of theoretical results shown in Proposition 1. Therefore, CAT is a generic method complementary to the existing point cloud processing techniques.
\setlength{\tabcolsep}{5pt} 
\begin{table}
	\begin{center}
		\caption{Ablation Study. All the experiments are implemented on ModelNet40\cite{wu20153d} against arbitrary rotations.}
		\label{ablation}
		\begin{tabular}{c|ccc|c}
			\toprule
			Method     & \#points  & $CAT$ & $FA$ & Acc(\%)\\
			\hline
			A & 1k    &  & & 20.6 \\
			B & 1k    &  & $\surd$ & 87.9 \\
			C & 1k    & $\surd$ &  & 89.0 \\
			D & 1k    & $\surd$ & $\surd$ & 89.5 \\
			E & 2k    & $\surd$ & $\surd$ & 89.5 \\
			\bottomrule
		\end{tabular}
	\end{center}
\vspace{-0.4cm}
\end{table}

\vspace{-0.2cm}
\subsection{Ablation Study} In this section, we conduct an ablation study of SCT to figure out how each module affects the overall performance. All experiments are implemented on ModelNet40\cite{wu20153d} against rotations, and the performance metric is the accuracy (\%). Models are trained without rotation augmentation and tested under arbitrary rotations. Table \ref{ablation} illustrates the results of the ablation study with $(x,y,z)$ coordinates.
The baseline (Model A) denotes DGCNN\cite{wang2019dynamic}, only remaining a classification accuracy of 20.6\%. When employed with FA, it obtains rotation invariance and is improved to 87.9\% (model B). The interpretation is that FA transforms each object into an intra-class frame. As rotation perturbations could be viewed as frame transformations, each rotated object has a different frame. Therefore, FA adjusts frames of rotated objects, and accordingly, model C harvests rotation robustness. Besides, we apply the CAT with DGCNN (model C), which outperforms FA methods. Moreover, combining the FA and CAT (model D) gives model C another 0.5\% boost and achieves state-of-the-art under arbitrary rotations.

Based on the performances of models A, B, C, and D, it is evident that both CAT and FA improve rotation robustness. We train the model with 2,048 points (Model E) but find no boost.

\vspace{-0.2cm}
\subsection{Robustness Evaluation}
In this section, we conduct experiments to evaluate the robustness of the proposed method under different sampling densities and noises. To this end, the robustness of RTIT on sampling density is shown in Fig. \ref{fig:sample1}. We test the CAT with sparser points of 1024, 768, 512, and 256, respectively. For a fair comparison, all the RTIT methods are fed into DGCNN model. Note that we do not use random input dropout augmentation during training. Visualization results in the leftmost of Fig. \ref{fig:sample1} demonstrate that CAT consistently transforms points based on contours under different sampling densities. Fig. \ref{fig:sample1} shows that our CAT is more robust than other methods. The interpretation is that nonlinear transformation techniques rely on local geometry structures, which are more sensitive to density differences. Besides, the Gaussian noise with different standard deviation (Std) is added to each point independently, shown in the right of Fig. \ref{fig:sample1}. CAT is more robust than other methods in terms of noise perturbations. Moreover, FA module further improves the robustness of CAT and adaptively transforms point clouds.

\begin{figure}[t]
	\vspace{-0.3cm}
	\centering
	\includegraphics[width=0.95\linewidth]{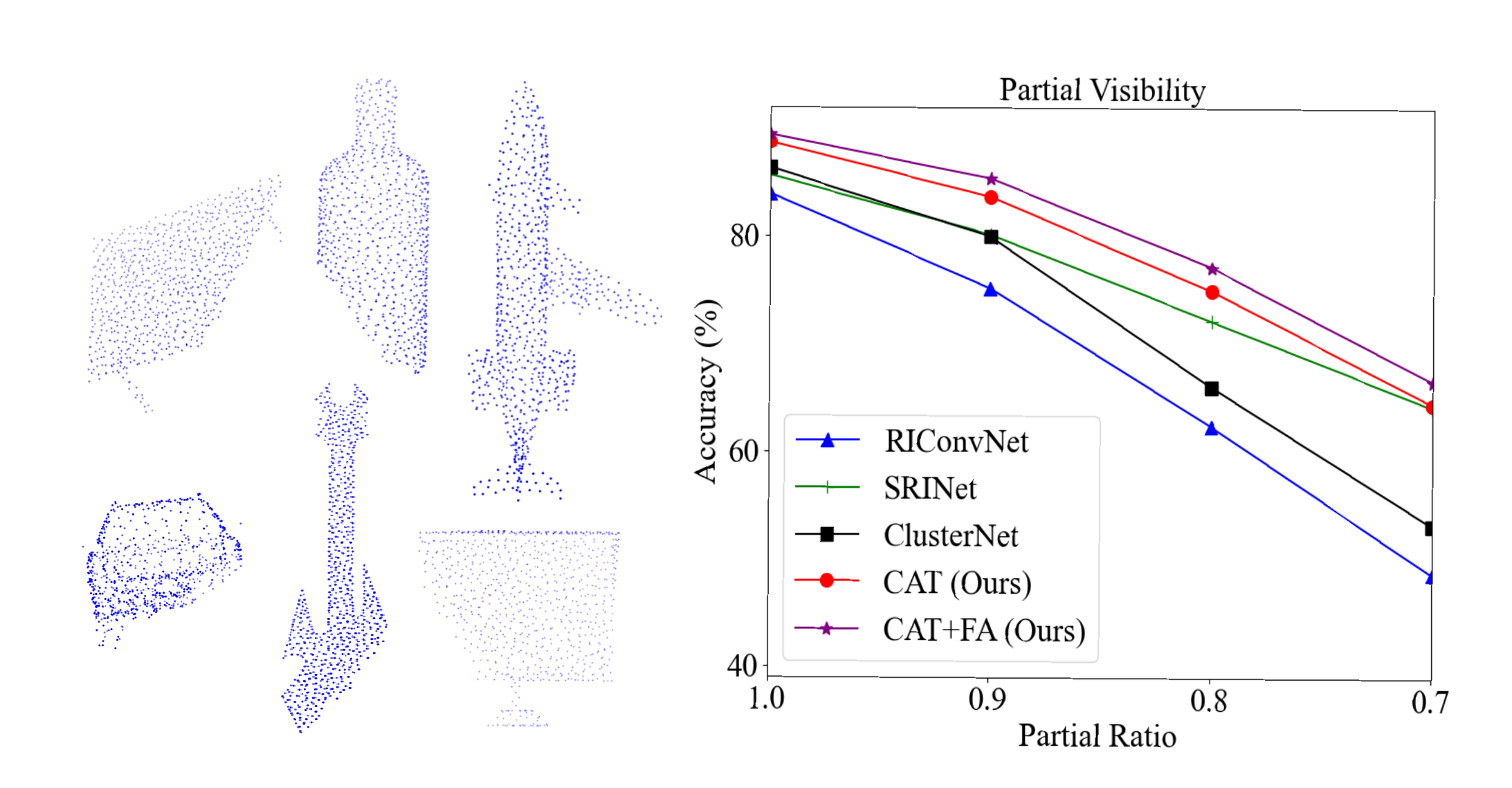}
	\caption{Evaluation on partial visible point clouds. All methods are evaluated in NR/AR.}
	\label{fig:partial}
	\vspace{-0.2cm}
\end{figure}

\subsection{Partial Visibility}
In this section, we conduct experiments on partially visible point clouds, which are common in autonomous driving applications. We simulate partial visibility following \cite{yew2020rpm} with partial ratios from 1 to 0.7. Specifically, all compared models are trained on the complete point cloud (1024 points). The leftmost of Fig. \ref{fig:partial} demonstrates some visualization examples of partially visible point clouds. The rightmost of Fig. \ref{fig:partial} shows that the proposed CAT is more robust than other methods. When nonlinear RTIT techniques handle partially visible point clouds, local graphs generated from complete and partial point clouds are different. Therefore, the proposed CAT is more suitable for real-world applications with partial visibility point clouds.

\subsection{Model Complexity}
 We conduct comparisons of model complexity, inference time and performances against rotations on ModelNet40 \cite{wu20153d} in Table \ref{inference_time}. For a fair comparison, we adapt DGCNN as the backbone for rotation-variant methods. We report the inference time with batch size 16. Table \ref{inference_time} shows that CAT achieves the best computational complexity (measured as inference time) among these rotation-invariant methods. Equipping CAT and FA improves the classification accuracy of DGCNN from 20.6\% to 89.5\% against rotations. This improvement is at the price of 0.8M parameters and a 5.9 ms inference time increase. Therefore, the proposed method achieves the best trade-off between accuracy and computational complexity and is applicable to real-time industrial scenarios.
 
 \renewcommand\arraystretch{1} 
\setlength{\tabcolsep}{3pt} 
\begin{table}
	\begin{center}
		\caption{\textbf{Comparisons of model complexity on ModelNet40.} $^{*}$ denotes our methods.}
		\label{inference_time}
		\begin{tabular}{cccc}
			\toprule
			Method     & Model size & Inference time  & NR/AR (\%) \\
			\midrule
			DGCNN  & 1.81M & \bf 28.4 ms  & 20.6 \\
			SRINet\cite{sun2019srinet} & 1.81M & 29.6 ms & 86.1 \\
			RI-Conv\cite{li2020rotation}  & 1.83M & 40.3 ms & 82.6 \\
			ClusterNet\cite{chen2019clusternet} & 1.85M & 42.8 ms & 86.4  \\
			CAT$^{*}$  & 1.81M & 29.1 ms  & 89.0 \\
			CAT$^{*}$+ FA$^{*}$ & 2.61M & 34.3 ms  &\bf 89.5 \\
			\bottomrule
		\end{tabular}
	\end{center}
\vspace{-0.5cm}
\end{table}

\section{Conclusion}
In this paper, a novel SCT has been proposed for point cloud recognition against arbitrary rotations. SCT provides efficient and strict rotation and translation invariance by introducing CAT. CAT transforms Cartesian coordinates of points to rotation and translation-invariant representations. Moreover, a FA module has been proposed to transform point clouds to the intra-class frame to enhance discriminative shape awareness extraction.
\par Furthermore, as demonstrated in theoretical analysis and experimental results, CAT outperforms nonlinear methods with lower computational complexity. CAT does not rely on local geometry structures of point cloud and is more robust under density differences and noises. More crucially, SCT can be flexibly integrated into existing point cloud processing models. Experimental results demonstrate that SCT achieves state-of-the-art performance on classification and part segmentation tasks on synthetic and real-world datasets against arbitrary rotations. In summary, owing to its efficiency, robustness under noises, sample densities, and partial visibility, and generality, SCT can be applied in real-world industrial applications.

\appendix
\setcounter{equation}{0}
\renewcommand\theequation{A.\arabic{equation}}
For ${\bm x}=(x_{1},x_{2},x_{3})^{T} $ and ${\bm y}=(y_{1},y_{2},y_{3})^{T} $, the inner product is given by,
\begin{equation}
\bm x\cdot \bm y = {\bm x^{T}\bm y}={\bm y^{T}\bm x}.
\label{A1}
\end{equation}
It is convenient to express matrix in terms of its columns. For a 3x3 matrix $M\equiv[{\bm {a\ b\ c}}]$, the adjoint\cite{markley2014fundamentals} is
\begin{equation}
\begin{aligned}
{\rm adj} M = {\rm adj}([{\bm {a\ b\ c}}])=\left[ \begin{array}{c}
{ (\bm b\times \bm c)^{T}} \\
{(\bm c\times \bm a)^{T}} \\
{(\bm a\times \bm b)^{T}} 
\end{array} 
\right ]
\end{aligned}
\label{A2}
\end{equation}
And $[\bm x \times]$ is the cross product matrix, defined by
\begin{equation}
[\bm x \times]={\left[ \begin{array}{ccc}
	0 & -x_{3} & x_{2}\\
	x_{3} & 0 & -x_{1}\\
	-x_{2} & x_{1} & 0
	\end{array} 
	\right ]}
\label{A3}
\end{equation}
Then
\begin{equation}
\begin{aligned}
M^T[{\bm x}\times]M = {
	\left[ \begin{array}{ccc}
	{\bm a^{T} (\bm x\times \bm a)} & {\bm a^{T}(\bm x\times \bm b)} & {\bm a^{T}(\bm x\times \bm c)}\\
	{\bm b^{T} (\bm x\times \bm a)} & {\bm b^{T}(\bm x\times \bm b)} & {\bm b^{T}(\bm x\times \bm c)}\\
	{\bm c^{T} (\bm x\times \bm a)} & {\bm c^{T}(\bm x\times \bm b)} & {\bm c^{T}(\bm x\times \bm c)}
	\end{array} 
	\right ]}
\end{aligned}
\label{A4}
\end{equation}
According to Eq. (\ref{A1}) , Eq. (\ref{A4}) is equal to

\begin{equation}
\begin{aligned}
M^T[{\bm x}\times]M \\=&  {
	\left[ \begin{array}{ccc}
	{\bm{a\cdot(x\times a)}} & {\bm{a\cdot(x\times b)}} & {\bm{a\cdot(x\times c)}}\\
	{\bm{b\cdot(x\times a)}} & {\bm{b\cdot(x\times b)}} & {\bm{b\cdot(x\times c)}}\\
	{\bm{c\cdot(x\times a)}} & {\bm{c\cdot(x\times b)}} & {\bm{c\cdot(x\times c)}}
	\end{array} 
	\right ]} \\
=& {
	\left[ \begin{array}{ccc}
	0 & {\bm{-(a\times b)\cdot x)}} & {\bm{(c\times a)\cdot x}}\\
	{\bm{(a\times b)\cdot x}} & 0& {\bm{-(b\times c)\cdot x}}\\
	{\bm{-(c\times a)\cdot x}} & {\bm{(b\times c)\cdot x}} & 0
	\end{array} 
	\right ]}\\
\end{aligned}
\label{A5}
\end{equation}

Considering Eq. (\ref{A2}), and Eq. (\ref{A3}), Eq. (\ref{A5}) reduces to
\begin{equation}
M^T[{\bm x}\times]M = [\{({\rm adj} M){\bm x}\}\times].
\end{equation}

Setting $M = R^{T}$, when $R$ is a proper orthogonal 3x3 matrix and adjacency matrix ${\rm adj}M=R$, then
\begin{equation}
R[{\rm \mathbf x}\times]R^{T} = [(R{\rm \mathbf x })\times] , for R \in SO(3).
\end{equation}

\ifCLASSOPTIONcaptionsoff
  \newpage
\fi



%
\bibliographystyle{IEEEtran}

%




\end{document}